\documentclass[10pt,journal,compsoc,10pt]{IEEEtran}
\usepackage[latin9]{inputenc}
\usepackage{color}
\usepackage{array}
\usepackage{multirow}
\usepackage{amsmath}
\usepackage{graphicx}

\makeatletter

\providecommand{\tabularnewline}{\\}

\usepackage{algorithm}
\usepackage{algorithmic}
\usepackage{subcaption}
\usepackage{multirow}
\usepackage{amsopn}

\DeclareMathOperator*{\argmax}{arg\,max}

\makeatother

\begin{document}
\title{Learning to Generate Posters of Scientific Papers by Probabilistic Graphical Models}
\author{Yuting Qiang,         Yanwei Fu, Xiao Yu, Yanwen Guo, \\ Zhi-Hua Zhou and Leonid Sigal
\IEEEcompsocitemizethanks{ \IEEEcompsocthanksitem Yuting Qiang, Xiao Yu, Yanwen Guo and Zhi-Hua Zhou are with National Key Laboratory for Novel Software Technology, Nanjing University Nanjing 210023, China. Email: qiangyuting.new@gmail.com, mg1533099@smail.nju.edu.cn, \{ywguo,zhouzh\}@nju.edu.cn.\protect\\ 
Yanwei Fu is with the school of Data Science, Fudan University, Shanghai, China. Email: yanweifu@fudan.edu.cn \protect\\  Leonid Sigal is with Disney Research, USA. Email: lsigal@disneyresearch.com 
}
} 
\IEEEtitleabstractindextext{
\begin{abstract} 
Researchers often summarize their work in the form of scientific posters. Posters provide a coherent and efficient way to convey core ideas expressed in scientific papers. Generating a good scientific poster, however, is a complex and time consuming cognitive task, since such posters need to be {\emph{readable}}, {\emph{informative}}, and visually {\emph{aesthetic}}. In this paper, for the first time, we study the challenging problem of learning to generate posters from scientific papers. To this end, a data-driven framework, that utilizes graphical models, is proposed. Specifically, given content to display, the key elements of a good poster, including attributes of each panel and arrangements of graphical elements are learned and inferred from data. During the inference stage, an MAP inference framework is employed to incorporate some design principles. In order to bridge the gap between panel attributes and the composition within each panel, we also propose a recursive page splitting algorithm to generate the panel layout for a poster. To learn and validate our model, we collect and release a new benchmark dataset, called NJU-Fudan Paper-Poster dataset, which consists of scientific papers and corresponding posters with exhaustively labelled panels and attributes. Qualitative and quantitative results indicate the effectiveness of our approach. \end{abstract} 
  } \maketitle

\section{Introduction}

\label{sec:introduction}

The emergence of a large number of scientific papers in various academic
fields and venues (conferences and journals) is noteworthy. For example,
ArXiv, a premiere on-line scientific repository, reports upload rate
of over 9,000 papers and reports {\em a month} in 2016. It is time-consuming
to read and digest all of these papers for researchers, particularly
those interested in holistically assess state-of-the-art, or understanding
of just core scientific ideas explored in the last year. Converting
a scientific paper into a poster provides an important way to efficiently
and coherently convey core ideas and findings of the original paper.

To achieve this goal, it is therefore essential to keep the posters
\emph{readable}, \emph{informative} and \emph{visually aesthetic}.
It is challenging, however, to design a high-quality scientific poster
which meets all of the above design principles, particularly for novice
researchers who may not be proficient at design tasks or familiar
with design packages (e.g., Adobe Illustrator). In general, poster
design is a complicated and time-consuming task; both understanding
of the paper content and experience in design are required.

Automatic tools for scientific poster generation would help researchers
by providing them with an easier way to effectively share their research.
Further, given avid amount of scientific papers on ArXiv and other
on-line repositories, such tools may also provide a way for other
researchers to consume the content more easily. Rather than browsing
raw papers, they may be able to browse automatically generated poster
previews (potentially constructed with their specific preferences
in mind).

Page layout generation \cite{jahanian2012automatic,Hunter2011SPIE,odonovan2014},
has been popular in recent years with the goal of generating graphical
design layout, such as photo collage\cite{Geigel03Mul}, furniture
object arrangements\cite{craigyu2011furniture,Merrell11ACM}, comics
panel layouts \cite{Cao2014Siggraph} and so on. These works pay more
attention on \emph{visual} \emph{aesthetics} than informativeness
and readability. On the other hand, there are also lots of works that
study presentation layout automation \cite{Hurst2009,knuth1981breaking,Peels1985ACM},
which aim at document generation. These works often focus on micro-typography
problems such as line breaking, margins inference and so on. In addition,
some works utilize templates as input to their layout algorithms \cite{Damera2011}.

In general, in order to generate a scientific poster in accordance
with, and representative of, the original paper, many problems need
to be solved: (1)\textbf{Content~extraction}. Both important textual
and graphical content needs to be extracted from the original paper;
(2)\textbf{ Panel~layout}. Content should fit each panel; and the
shape and position of panels should be optimized for readability and
design appeal; (3)\textbf{ Graphical~element~(figure~and~table)~arrangement}.
Within each panel, textual content can typically be sequentially presented,
but for graphical elements, their size and placement should be carefully
considered. Due to these challenges, to our knowledge, no automatic
tools for scientific poster generation exist.

In this paper, we propose a data-driven method for automatic scientific
poster generation (given a corresponding paper). Content extraction
and layout generation are two key components in this process. For
content extraction, we use TextRank \cite{MihalceaT04emnlp} to extract
textual content, and provide an interface for extraction of graphical
content (e.g., figures, tables, etc.). Our approach focuses primarily
on poster layout generation. We address the layout in three steps.
First, we propose a probabilistic graphical model to infer panel attributes.
Second, we introduce a tree structure to represent panel layout, based
on which we further design a recursive algorithm to generate new layouts.
Third, in order to synthesize layout within each panel, we train another
probabilistic graphical model to infer the attributes of graphical
elements.

Compared with posters designed by the authors, our approach is more
efficient and versatile. Our approach can generate results that adapt
to different paper sizes/aspect ratios or styles, by training our
model with different dataset. 

To the best of our knowledge, this paper presents the first method
for scientific poster generation from the original academic papers.
A preliminary version of this work appeared as a conference paper
\cite{Qiang2016AAAI}. This paper extends the previous version in
the following perspectives: (1)\textbf{\textcolor{black}{Enlarged~dataset.}}
We have enlarged and released our dataset\footnote{see http://www.ytqiang.com/}
to the community 
as a new benchmark dataset for evaluating the problem of scientific
poster generation. (2)\textbf{\textcolor{black}{Improved~methodology.}}
We improve our method in several ways: (1) we propose a novel loss
function to evaluate the panel arrangement, which helps our algorithm
to find better panel layouts. (2) We refine the probabilistic graphical
model framework for element composition within each panel, this refinement
takes some design principles into consideration and makes our approach
more effective. (3)\textbf{\textcolor{black}{Additional~Experiments.}}
We provide more detailed performance analysis and extensive experiments
to show the effectiveness of the new method. 

The remainder of this paper is organized as follows. The related works
is briefly introduced in Section 2. In Section 3, we describe our
dataset and preprocessing work in detail. In section 4 and 5, we present
a high-level overview and key components of our method separately.
Experiments and evaluation are discussed in Section 6. 

\section{Related Work}

\label{sec:related_works}

In this section, we review three heavily studied topics of page layout
generation, \emph{i.e.}, general graphical design (Sec. \ref{subsec:General-Graphical-Design}),
comic layout generation (Sec. \ref{subsec:Comic-Layout-Generation})
and presentation layout automation (Sec. \ref{subsec:Presentation-Layout-Automation}),
and the differences between these topics and our task of scientific
poster generation. 

\subsection{General Graphical Design\label{subsec:General-Graphical-Design}}

Graphical design has been studied extensively in computer graphics
community. This involves several related, yet different topics. Geigel
\emph{et al.} \cite{Geigel03Mul} made use of genetic algorithm \cite{Holland1992ANA,Goldberg1989GAS}
for \emph{photo album layout}, which addresses the placement of each
photo in an album. Yu \emph{et al.} \cite{craigyu2011furniture} automatically
synthesized furniture objects arrangements using simulated annealing
algorithm. In contrast, Merrell \emph{et al.} \cite{Merrell11ACM}
applied some simple design guidelines to solve a similar problem.
Other graphical design problems such as \emph{interface design} \cite{Gajos05UIST},
\emph{circuit board layout} \cite{Sarrafzadeh1993AAV}, and \emph{graph
layout} \cite{Battista1998GDA} have also been studied. These works
often present an optimization framework along with some design guidelines
to synthesize and evaluate plausible layouts.

Nevertheless, these works are concerned more about graphical elements
(e.g., photo, furniture), and they take visual aesthetics as the highest
priority. In contrast, for scientific poster generation, textual content,
original paper structure and the order of contents need to be considered
to ensure the readability of a scientific poster.

\subsection{Comic Layout Generation\label{subsec:Comic-Layout-Generation}}

Due to the popularity of comics, many related research topics, such
as \emph{manga retargeting} \cite{Matsui2011SIG}, \emph{comic episodes
generation} \cite{Hoashi2011MM} and \emph{manga-like rendering} \cite{Qu2008ACM}
have drawn considerable research attention in computer graphics community.
Particularly, several techniques have been studied to facilitate layout
generation. For example, Arai \emph{et al.} \cite{arai2010method}
and Pang \emph{et al.} \cite{Pang2014MM} studied how to automatically
extract each panel from e-comics; and display e-comics on different
devices. In order to convert conversational videos to comics, Jing
\emph{et al.} \cite{Jing15TMM} made use of a rule based optimization
scheme for layout generation. Cao \emph{et al.} \cite{Cao2012TOG}
presented a generative probabilistic framework to arrange input artworks
into a manga page, and then used optimization techniques to refine
it. Furthermore, Cao \emph{et al.} \cite{Cao2014Siggraph} took text
balloons and picture subjects into consideration for manga layout
generation and guided the reader's attention. However, in our poster
generation, one has to consider both texts and graphical elements
composition within each panel, which has not been discussed previously.

Our panel layout generation method is partly inspired by the recent
work on manga layout \cite{Cao2012TOG}. We use a binary tree to represent
the panel layout. By contrast, the manga layout trains a Dirichlet
distribution to sample a splitting configuration, and different Dirichlet
distribution for each kind of instance has to be trained as a result.
Instead, we propose a recursive algorithm to search for the best splitting
configuration along a binary tree.

\subsection{Presentation Layout Automation\label{subsec:Presentation-Layout-Automation}}

The emergence of data and information that we need to present, challenges
our ability to present it manually; thus, automated layout of presentations
is becoming increasingly important \cite{Hurst2009}. For \emph{automated
document formatting}, early works, such as \cite{knuth1981breaking,Peels1985ACM},
focused largely on line breaking, paragraph arrangement and some other
micro-typography problems. A common way to solve these problems is
as a constrained optimization problem \cite{lok2001survey}. More
recent works pay attention to \emph{presentation document layout}.
Jacobs \emph{et al.} \cite{Jacobs2003ACM} presented a grid based
dynamic programming method to select a page layout template. Damera-Venkata
\emph{et al.} \cite{Damera2011} made use of Probabilistic Document
Model (PDM) to facilitate document layout. By contrast, we focus on
both macro-typography problems (e.g panel layout) and micro-typograph
(e.g. graphical elements size decision) in this paper. Additionally,
rather than using simple design guidelines as previous work \cite{knuth1981breaking,Peels1985ACM},
we learn our layout generating model from the annotated training datasets.

Another piece of relate work is called\emph{ single page graphical
design} \cite{odonovan2014}, which made use of an energy-based model
derived from design principles for graphic design layout. However,
they regard texts as a rectangle block rather than text flow, which
is inappropriate for scientific poster generation. Harrington \emph{et
al.} \cite{harrington2004aesthetic} described a measure of document
aesthetics, and an aesthetics driven layout engine is proposed in
\cite{Purvis2003CPD}. However, these approaches do not put constraints
on the ordering of content, which is clearly important for scientific
poster generation. 

\section{The NJU-Fudan Paper-Poster Da-\protect \\
 taset}

\label{sec:NJU-Fudan_dataset}

\begin{figure}
\centering{}\includegraphics[width=0.35\textwidth]{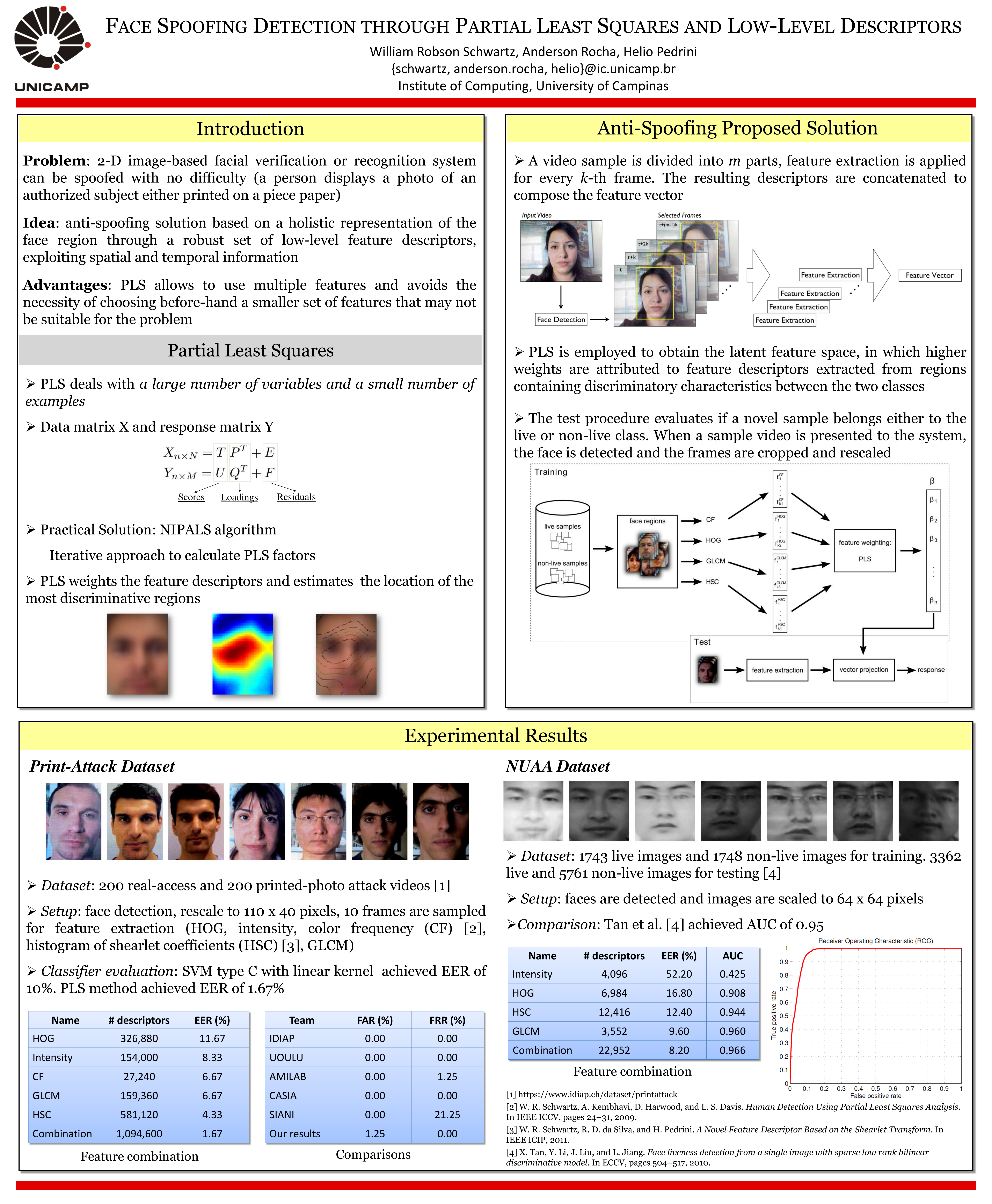} \caption{\label{fig:example} An example of human designed poster.}
\end{figure}

In this paper, we propose a new research topic of learning to generate
posters of scientific papers. In general, a good poster for a scientific
paper should follow the general design principles. One section in
the paper should correspond to one panel in the poster. Each panel
usually includes several bullet points and sentences that explain
the corresponding bullet point, and each bullet point often corresponds
to a sub-section or a paragraph in the paper. Important figures and
tables in each paper section would also be included in the corresponding
poster panel. Figure \ref{fig:example} shows such an example of human
designed poster \cite{Pinto2015TIP}. This type of scientific poster
is \emph{readable}, \emph{informative} and \emph{visually aesthetic}
since it considers both the structure and key messages conveyed by
the original paper, which makes it easy for readers to understand.

To further study the tasks of poster generation for scientific papers,
we introduce a NJU-Fudan Paper-Poster dataset which contains pairs
of scientific posters and their corresponding papers. A total of 85
computer science research paper-poster pairs were collected from an
online website. 

We further annotate the meta information for each paper-poster to
facilitate the research of this topic. For each poster, we label both
layout attributes (e.g. panel position, figure size) and content attributes
(e.g. text length in each panel). In the corresponding paper, layout
related information (e.g. figure size in original paper) is also manually
labelled. We also provide annotation tool which can enable the annotation
and labeling of further data. 
Both the dataset and annotation tool will be released.

\section{Overview}

\label{sec:Overview}

\begin{figure}
\centering{}\includegraphics[width=0.5\textwidth]{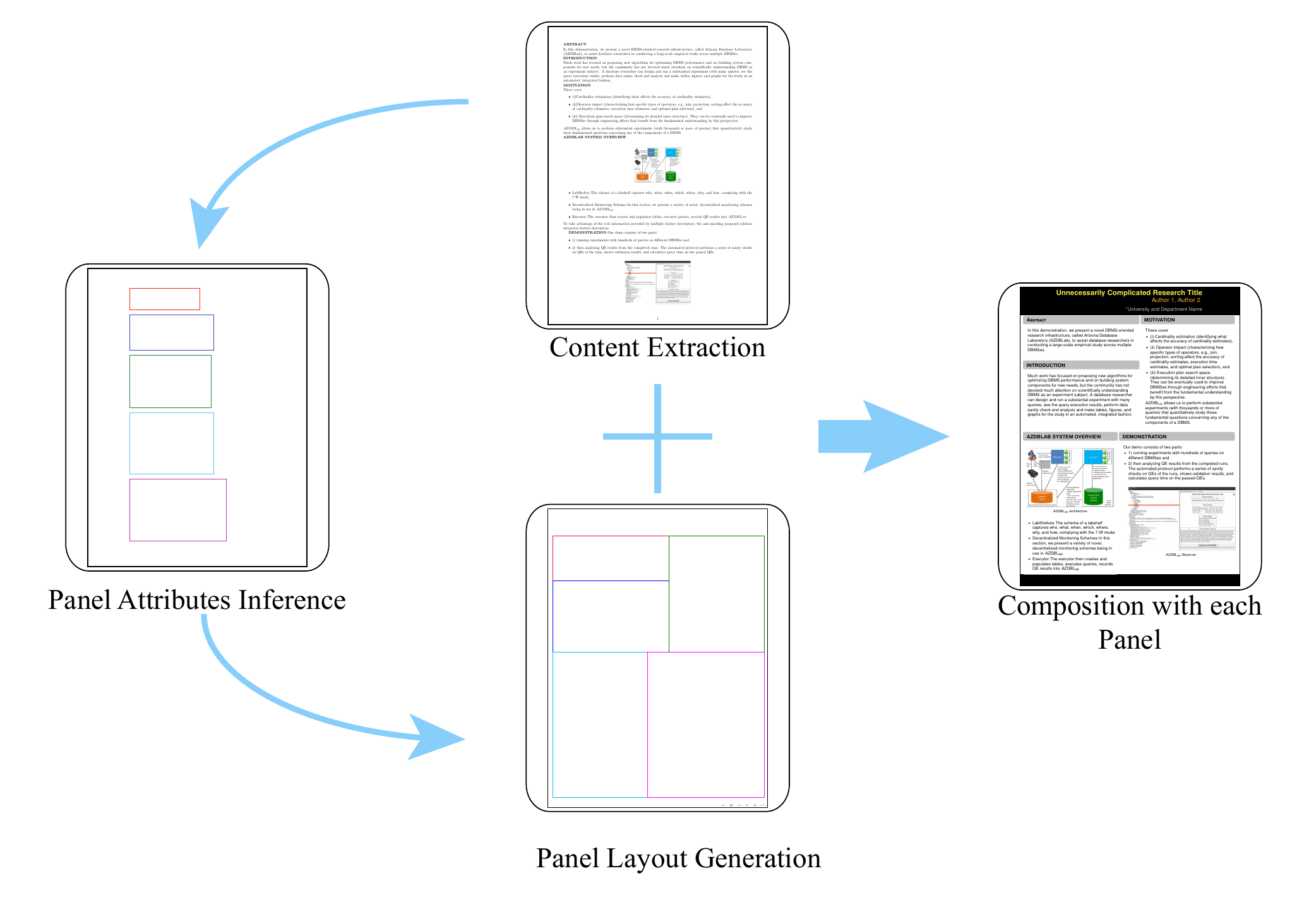}
\caption{ Overview of the proposed approach.}
\label{fig:Overview-of-our} 
\end{figure}

\noindent \textbf{Overview.} To generate a \emph{readable}, \emph{informative}
and \emph{aesthetic} poster, we simulate the rule-of-thumb on how
the researchers design posters in practice. We generate the panel
layout for a scientific poster first, and then arrange the textual
and graphical elements within each panel. As shown in Figure~\ref{fig:Overview-of-our},
the framework overall has four steps, namely, \emph{content extraction},
\emph{panel attributes inference}, \emph{panel layout generation},
and \emph{composition within each panel}.

\vspace{0.1in}

\noindent \textbf{Problem Formulation.} We formally introduce the
problem of learning to generate posters of scientific papers before
developing our contributions to each section. We have a set of posters
\textbf{M} and their corresponding scientific papers. Each poster
$m\in\mathbf{M}$ includes a set of panels $\mathbf{P}_{m}$, and
each panel $p\in\mathbf{P}_{m}$ has a set of graphical elements (figures
and tables) $\mathbf{G}_{p}$. Each panel $p$ is characterized by
six attributes: 
\begin{itemize}
\item text length within a panel ($l_{p}$); 
\item text~ratio~($t_{p}$), text length within a panel relative to text
length of the whole poster, $t_{p}=l_{p}/\sum_{q\in\mathbf{P}_{m}}l_{q}$; 
\item number of graphical elements within a panel ($n_{p}$); 
\item graphical~elements~ratio~($g_{p}$), the size of graphical elements
within a panel relative to the total size of graphical elements in
the poster. Note that there is a little difference between $g_{p}$
and $t_{p}$. Here instead of predicting the fixed figure size in
poster, we directly use the corresponding figure from original paper; 
\item panel~size~($s_{p}$)~and~aspect~ratio~($r_{p}$), $s_{p}=w_{p}\times h_{p}$
and $r_{p}=w_{p}/h_{p}$, where ${w_{p}}$ and ${h_{p}}$ denote the
width and height of a panel with respect to the poster.
\end{itemize}
Each graphical element $g\in\mathbf{G}_{p}$ has four attributes: 
\begin{itemize}
\item graphical~element~size~($s_{g}$) and aspect ratio ($r_{g}$),
$s_{g}=w_{g}\times h_{g}$ and $r_{g}=w_{g}/h_{g}$, where ${w_{g}}$
and ${h_{g}}$ denote the width and height of a graphical element
relative to the whole paper respectively; 
\item horizontal~position~($h_{g}$), inspired by the way how latex beamer
makes poster, we arrange that panel content sequentially from top
to bottom; hence only relative horizontal position needs to be considered,
which is defined by a discrete variable $h_{g}\in\{left,center,right\}$; 
\item graphical~element~size~in~poster~($u_{g}$), the ratio of the
width of the graphical element with width of the panel it belongs
to. 
\end{itemize}
To learn how to generate a poster, our goal is to determine the above
attributes for each panel $p\in\mathbf{P}_{m}$ and each graphical
element $g\in\mathbf{G}_{p}$, as well as the arrangement of the panels.

Intuitively, a trivial solution is to use a learning model (e.g.,
Support Vector Regression (SVR)) to learn how to regress these attributes,
including $s_{p}$, $r_{p}$, $u_{g}$, and $h_{g}$, while regarding
attributes which can be known according to corresponding scientific
paper (i.e. $t_{p}$, $g_{p}$, $l_{p}$,$r_{g}$, and $s_{g}$) as
features. However, such a solution takes those features as a whole,
thereby lacks an insight mechanism for exploring the relationships
between specific attributes (e.g. $s_{p}$ and $g_{p}$). It may fail
to meet the requirements of readability, informativeness and aesthetics.
We thus propose a Bayesian network to characterize the relationships
among those attributes, where the Bayesian network is trained on the
paper-poster dataset we collected. Then according to the Bayesian
network we trained, we can infer the layout attributes by using likelihood-weighted
sampling method.

\section{Methodology}

\label{sec:methodology}

In this section, we will further explain each step of our framework
as illustrated in Figure~\ref{fig:Overview-of-our}. Particularly,
(1) in Sec. \ref{subsec:Content-Extraction} we extract from the paper
the text content and graphical content. The textual content can be
summarized by the textual summary algorithms; and the graphical content
(figures and tables) would usually occupy a rectangular area of the
poster, and be extracted by user interactions. All extracted contents
are sequentially arranged. (2) Inference of the key attributes for
initial panel (such as panel size $s_{p}$ and aspect ratio $r_{p}$)
is then conducted by learning a probabilistic graphical model from
the training data in Sec. \ref{subsec:Panel-Attribute-Inference}.
(3) Furthermore, Sec. \ref{subsec:Panel-Layout-Generation} synthesizes
\emph{panel layout} by developing our recursive algorithm to further
update these key attributes (i.e., $r_{p}$) and generate an \emph{informative}
and \emph{aesthetic} panel layout. (4) Finally, we compose these panels
by utilizing our graphical algorithm to further synthesize the visual
properties of each panel (such as the size and position of its graphical
elements) in Sec. \ref{subsec:Composition-within-a}.

\subsection{Content Extraction \label{subsec:Content-Extraction}}

Content extraction, which includes both textual content extraction
and graphical content extraction, is the first step in our proposed
scientific poster generation system.

For textual content, we employ the state-of-the-art textual summary
algorithm to summarize the content of each section. In particular,
we use TextRank~\cite{MihalceaT04emnlp}. 

For graphical content, our algorithm will parse the key meta data
of the layout (i.e. width and height) of each figure and table. To
better select the most important figures/tables, we add user interaction
here to rank the importance of the tables and figures.

\subsection{Panel Attributes Inference\label{subsec:Panel-Attribute-Inference}}

We assume that in the poster each section should be represented by
one rectangular panel, which should not only be of an appropriate
size to contain the textual and graphical content of each corresponding
section, but also be in a reasonable shape (aspect ratio) to maximize
visually aesthetic appearance.

To enable such a goal, we learn a Bayesian network to infer the initial
size and aspect ratio for each panel. As each panel is composed of
both textual description and graphical elements, we assume that panel
size ($s_{p}$) and aspect ratio ($r_{p}$) are conditionally dependent
on text ratio $t_{p}$, number of graphical elements $n_{p}$ and
graphical element ratio $g_{p}$. Therefore, we define the joint probability
of a set of panels $\mathbf{P}$ as, 
\begin{equation}
Pr(\mathbf{P}|T,N,G)=\prod_{p\in P}Pr(s_{p}|t_{p},n_{p},g_{p})Pr(r_{p}|t_{p},n_{p},g_{p})\label{eq:panel_infer}
\end{equation}
where $T=\{t_{p}|p\in\mathbf{P}\}$, $N=\{n_{p}|p\in\mathbf{P}\}$
and $G=\{g_{p}|p\in\mathbf{P}\}$ denote attributes set. $Pr(s_{p}|t_{p},n_{p},g_{p})$
and $Pr(r_{p}|t_{p},n_{p},g_{p})$ are conditional probability distributions
(CPDs) of $s_{p}$ and $r_{p}$ given $t_{p}$, $n_{p}$ and $g_{p}$.
We further model them as two conditional linear Gaussian distributions:
\begin{equation}
Pr(s_{p}|t_{p},n_{p},g_{p})=N({s_{p}};\mathbf{w_{s}}[t_{p},n_{p},g_{p},1]^{\mathsf{T}},\sigma_{s})
\end{equation}
\begin{equation}
Pr(r_{p}|t_{p},n_{p},g_{p})=N({r_{p}};\mathbf{w_{r}}[t_{p},n_{p},g_{p},1]^{\mathsf{T}},\sigma_{r})
\end{equation}
where $t_{p}$ and $g_{p}$ are defined by the \emph{content extraction}
step demonstrated in Figure \ref{fig:Overview-of-our}; $\mathbf{w_{s}}$
and $\mathbf{w_{r}}$ are parameters that leverage the influence of
various factors; $\sigma_{s}$ and $\sigma_{r}$ are the variances.
The parameters ($\mathbf{w_{s}}$, $\mathbf{w_{r}}$, $\sigma_{s}$
and $\sigma_{r}$) are estimated using maximum likelihood estimator.

Note that in order to learn from limited data, this step actually
employs two assumptions: (1) $s_{p}$ and $r_{p}$ are conditionally
independent; (2) the attribute sets for panels are independent.

\begin{algorithm}[htb] 
\caption{Panel layout generation} 
\label{alg:panel_gen} 
\begin{algorithmic}[1] 
\REQUIRE ~~\\Panels which we learned from graphical model\\ $L=\{(s_{p_1},r_{p_1}),(s_{p_2},r_{p_2}), \cdots,(s_{p_k},r_{p_k})\}$;\\ rectangular page area $x$, $y$, $w$, $h$. \\
\ENSURE ~~\\ \IF {$k==1$} 
\STATE adjust panels$[0]$ to adapt to the whole rectangular  area, return the aesthetic loss: $|r_{p_0}-w/h|$; 
\ELSE
\FOR {each $i\in[1,k-1]$} 
\STATE $t=\sum_{j=1}^{i} s_{p_j} / \sum_{j=1}^{n} s_{p_j}$; 
\STATE $Loss_1$ = Panel Arrangement($(s_{p_1},r_{p_1}),(s_{p_2},r_{p_2}),$ \\ $\cdots,(s_{p_i},r_{p_i})$, $x$, $y$, $w$, $h\times t$);
\STATE $Loss_2$ = Panel Arrangement($(s_{p_{i+1}},r_{p_{i+1}}),(s_{p_{i+2}},$\\ $r_{p_{i+2}}), \cdots,(s_{p_k},r_{p_k})$, $x$, $y+h\times t$, $w$, $h\times(1-t)$); \IF {$Loss > Loss_1+Loss_2+\alpha|t-0.5|$}
\STATE $Loss = Loss_1+Loss_2+\alpha|t-0.5|$; 
\STATE record this arrangement; 
\ENDIF 
\STATE $Loss_1$ = Panel Arrangement($(s_{p_1},r_{p_1}),(s_{p_2},r_{p_2}),$\\ $\cdots,(s_{p_i},r_{p_i})$, $x$, $y$, $w\times t$, $h$); 
\STATE $Loss_2$ = Panel Arrangement($(s_{p_{i+1}},r_{p_{i+1}}),(s_{p_{i+2}},$\\ $r_{p_{i+2}}),\cdots,(s_{p_k},r_{p_k})$, $x+w*t$, $y$, $w\times(1-t)$, $h$); 
\IF {$Loss > Loss_1+Loss_2+\alpha|t-0.5|$} 
\STATE $Loss = Loss_1+Loss_2+\alpha|t-0.5|$; 
\STATE record this arrangement; 
\ENDIF
\ENDFOR 
\ENDIF 
\RETURN Loss and arrangement. \end{algorithmic} \end{algorithm}

We need the panels to be neither too small in size ($s_{p}$), nor
too distorted in aspect ratio ($r_{p}$), to ensure a readable, informative
and aesthetic poster. The two assumptions introduced here are sufficient
for this task. Furthermore, the attribute values estimated in this
step are just good initial values for the property of each panel.
We use the next two steps to further relax these assumptions and discuss
the relationship between $s_{p}$ and $r_{p}$, as well as the relationship
among different panels (Algorithm \ref{alg:panel_gen}).

To ease exposition, we denote the set of panels as $P=\{(s_{p_{1}},r_{p_{2}}),(s_{p_{2}},r_{p_{2}}),\cdots,(s_{p_{k}},r_{p_{k}})\}$,
where $s_{p_{i}}$ and $r_{p_{i}}$ are the size and aspect ratio
of $i$-th panel $p_{i}$, separately; and $\left|P\right|=k$.

\begin{figure}
\centering \includegraphics[width=0.5\textwidth]{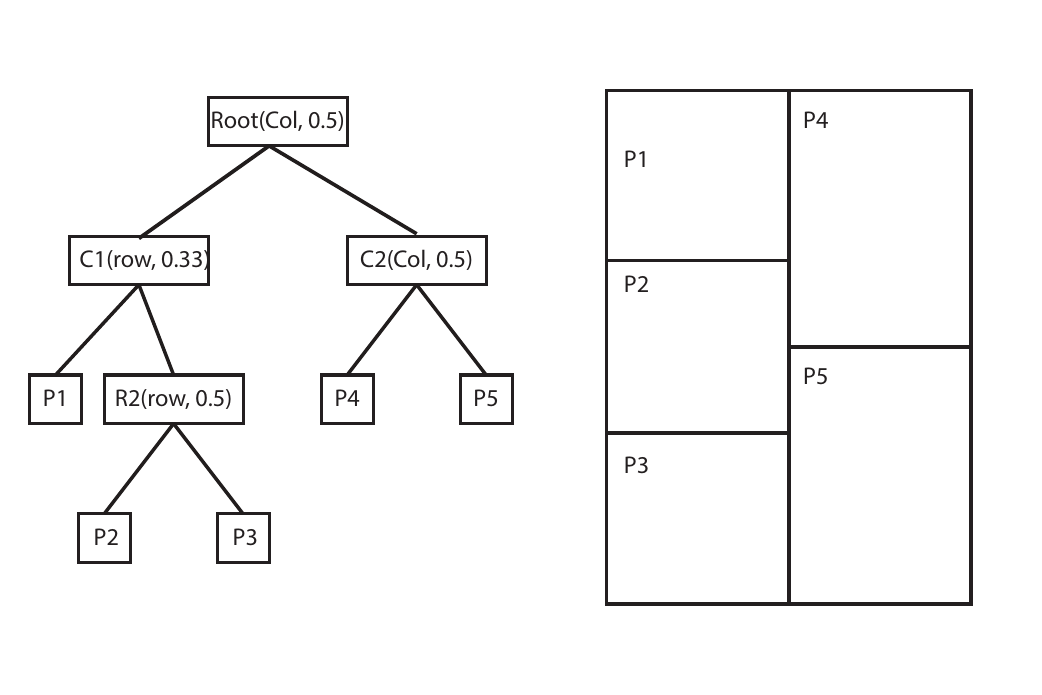}
\caption{\label{fig:tree_structure}Panel layout and the corresponding tree
structure. The tree structure of a poster layout contains five panels.
The first splitting is vertical with the splitting ratio (0.5, 0.5).
The poster is further divided into three panels in the left, and two
panels in the right. This makes the whole page as two equal columns.
For the left column, we resort to a horizontal splitting with the
splitting ratio (0.33, 0.67). The larger one is further horizontally
divided into two panels with the splitting ratio (0.5, 0.5). We only
split the right column once, with the splitting ratio (0.5, 0.5).}
\end{figure}

\subsection{Panel Layout Generation\label{subsec:Panel-Layout-Generation}}

One conventional way to design posters is to simply arrange them in
two or three columns style. This scheme, although simple, however,
makes posters designed in this way look similar. Inspired by manga
layout generation~\cite{Cao2012TOG}, we propose a more vivid panel
layout generation method. Specifically, we arrange the panels with
a binary tree structure to help represent the panel layout. The panel
layout generation is then formulated as a process of recursively splitting
of a page, as illustrated and explained in Figure \ref{fig:tree_structure}.

Conveying information is the most important goal for a scientific
poster, thus we attempt to maintain the relative size for each panel
during panel layout generation. This motivates the following loss
for the panel shape variation,

\begin{equation}
l_{var}(p_{i})=|r_{p_{i}}-r_{p_{i}}^{'}|\label{eq:panel_loss_function}
\end{equation}
where $r_{p_{i}}^{'}$ is the aspect ratio of a panel after optimization.

On the other hand, we also evaluate the aesthetic for the splitting
configuration. In our approach, the splitting configuration is composed
of several splittings. Each splitting divides a set of panels into
two parts, and the splitting ratio is decided by the ratio of the
total size of the two panels. Since symmetry is an important guideline
for design works, we evaluate the aesthetic for the panel layout configuration
based on the symmetry of each partition. In particular, if a panel
set is divided by a splitting $\zeta_{i}$ as ${p_{1},p_{2},\cdots,p_{k}}$
and ${p_{k+1},p_{k+2},\cdots,p_{m}}$, then the aesthetic loss for
this splitting is defined as follow: 
\begin{equation}
l_{aes}(\zeta_{i})=\alpha|\sum_{i=1}^{k}s_{p_{i}}/\sum_{i=1}^{m}s_{p_{i}}-0.5|\label{eq:splitting_loss_funtion}
\end{equation}

The loss for panel shape variation (Eq. \ref{eq:panel_loss_function})
and splitting configuration (Eq. \ref{eq:splitting_loss_funtion})
lead to a combined loss for the panel layout arrangement 
\begin{equation}
Loss(P,P^{'},Z)=\sum_{i=1}^{k}{l_{var}(p_{i})}+\sum_{\zeta\in Z}{l_{aes}(\zeta)}\label{eq:loss_function}
\end{equation}
where $P^{'}$ is the panel set after optimization and $Z$ is the
set of splitting steps.

In each splitting step, the combinatorial choices for splitting positions
can be recursively computed and compared with respect to the loss
function above. We choose the panel attributes with the lowest loss
(Eq. \ref{eq:loss_function}). The whole algorithm is summarized in
Algorithm \ref{alg:panel_gen}.

\subsection{Composition within a Panel\label{subsec:Composition-within-a}}

Having inferred the layout of panels, we turn our attention to the
composition of raw contents within each panel. Generally, each panel
in a scientific poster is composed of textual and graphical content.
Considering the readability of a scientific poster, each panel can
be filled by these contents sequentially. However, for aesthetic consideration,
the horizontal position and size of each graphical element need to
be specified carefully. Therefore, we pose automated panel composition
as an inference problem in a Bayesian network that incorporates some
design constraints. 

Designing the composition for each panel is complicated, both panel
attributes and raw contents need to be considered. We aim at designing
a Bayesian network to characterize how these variables interact with
each other. Given the placement of each graphical element, textual
contents can be filled into the panel sequentially; therefore, the
composition of a panel can be defined by the horizontal position ($h_{g}$)
and the size of each graphical element ($s_{g}$). In our approach,
the layout within each panel is composed by first sampling random
variable $h_{g}$ representing the choice of horizontal position (left,
right, center), and then sampling the variable $s_{g}$ representing
the size of a graphical element.

In our Bayesian network, horizontal position ($h_{g}$) of a graphical
element relies on both the shape ($r_{p}$) of the panel which the
element belongs to and attributes of the element ($r_{g}$, $s_{g}$)
itself. For example, a portrait figure is more likely to be presented
in the left or right of a landscape panel. To describe such relationship,
the horizontal position $h_{g}$ of a graphical element $g$ in panel
$p$ is sampled from a soft-max function, 
\begin{equation}
Pr(h_{g}=i|r_{p},r_{g},s_{g})=\frac{e^{\mathbf{w_{h}}_{i}\cdot[r_{p},r_{g},s_{g},1]^{\mathsf{T}}}}{\sum_{j=1}^{H}e^{\mathbf{w_{h}}_{j}\cdot[r_{p},r_{g},s_{g},1]^{\mathsf{T}}}}\label{eq:hpos_softmax_cpd}
\end{equation}
where $H=3$ is the cardinality of the value set of $h_{g}$; $\mathbf{w_{h}}_{i}$
is $i-$th row of $\mathbf{w_{h}}$.

For the size of a graphical element ($u_{g}$), it has to meet two
requirements: on the one hand, it needs to be appropriate to fill
the panel; on the other hand, it also needs to harmonize with the
occupation of the graphical element in the original paper. To this
end, in our model, the size of each graphical element ($u_{g}$) is
governed by both the panel attributes ($l_{p}$, $s_{p}$) and it's
own properties ($s_{g}$, $h_{g}$). We may sample the size of the
graphical elements from the conditional linear Gaussian distribution,
\begin{equation}
\begin{split}Pr(u_{g}|s_{p},l_{p},s_{g},h_{g})=\\
N({u_{g}} & |\mathbf{w_{u}}\cdot[s_{p},l_{p},s_{g},h_{g},1]^{\mathsf{T}},\mathbf{\sigma_{u}})
\end{split}
\label{eq:size_gaussian_cpd}
\end{equation}
where $\mathbf{w_{u}}$ is the parameter to balance the influence
of different factors.

For a set of graphical elements $G$ which belongs to the same panel
$p$, the probability of sampling process described above is simply
the product the probabilities of all design choice made during the
sampling process, it can be represented by the following distribution,
\begin{equation}
\begin{split}Pr(h_{G},u_{G}| & s_{p},r_{p},l_{p},s_{G},r_{G})=\\
\prod_{g\in G} & Pr(h_{g}|r_{p},r_{g},s_{g})Pr(u_{g}|s_{p},l_{p},s_{g},h_{g})
\end{split}
\ \label{eq:figure_likelihood_function}
\end{equation}

\noindent where $h_{G}$ and $u_{G}$ denote the assignments of horizontal
position and size in panel for all graphical elements in $G$, respectively;
$s_{G}$ and $r_{G}$ represent the input attributes of $G$.

\vspace{0.1in}

\noindent \textbf{Learning.} The goal of the learning stage in this
step is to estimate the parameters in our Bayesian network from training
data, this can be done by maximizing the complete-data log likelihood
since all the random variables in our model are observed. For conditional
linear gaussian distribution (Eq. \ref{eq:size_gaussian_cpd} ), with
some algebraic manipulation we can compute the optimal ML estimate
of $\mathbf{w_{u}}$ and $\mathbf{\sigma_{u}}$ in a closed form:
\begin{equation}
\begin{split}\mathbf{w_{u}^{*}} & =(\sum_{i}^{n}x^{(i)}x^{(i)T})^{-1}(\sum_{i}^{n}u_{g}^{i}x^{i})\\
\mathbf{\sigma_{u}^{*}} & =\frac{1}{n}\sum_{i=1}^{n}(u_{g}^{(i)}-\mathbf{w_{u}}^{*T}x^{(i)})^{2}
\end{split}
\end{equation}
where $x^{(i)}=[s_{p},l_{p},s_{g},h_{g},1]^{(i)}$ denote the training
data. For soft-max function (Eq. \ref{eq:hpos_softmax_cpd}), while
there is no known closed-form ML solution, we can resort to an iterative
optimization algorithm \textendash{} iteratively reweighted least
squares (IRLS) algorithm.

The Bayesian network described above models the relationship between
different variables explicitly. However, it is also desirable to consider
the relationship between panel size and content occupation. In a human
designed poster, contents usually fill each panel up exactly, which
makes the poster seems clean and informative. Therefore, we incorporate
the design principles with our Bayesian network, and our goal is to
find solution to this function: 
\begin{equation}
\begin{split}{h_{G}^{*},u_{G}^{*}} & =\argmax_{h_{G},u_{G}}f(h_{G},u_{G}|s_{p},r_{p},l_{p},s_{G},r_{G})\\
 & =\lambda_{1}\log Pr(h_{G},u_{G}|s_{p},r_{p},l_{p},s_{G},r_{G})\\
 & +\lambda_{2}\log N(w_{p}*h_{p}|\beta t_{p}+\sum_{g\in p}s_{g},\rho)
\end{split}
\label{eq:map_inference_function}
\end{equation}
in the equation above, the first term is defined in Eq. \ref{eq:figure_likelihood_function},
it is a likelihood that determines how well the solution fits our
Bayesian network. The second term measures how well the contents fit
the panel size, and it assigns high probability if the contents fill
the panel precisely and lower probability for deviations from the
ideal.

\noindent Since the exact MAP inference is not tractable in our model,
we perform approximate inference by using likelihood-weighted sampling
method~\cite{Murphy02}.

\section{Experimental Results}

\label{sec:expermental_result}

\subsection{Experimental Setup}

\noindent \textbf{NJU-Fudan Paper-Poster Dataset.} Our dataset includes
$85$ well-designed pairs of scientific papers and their corresponding
posters, which is selected from $600$ publicly available pairs we
collected. These papers are all about computer science topics, and
their posters have relatively similar design styles. We further annotate
panel attributes, such as panel width, panel height and so on. The
annotated meta data is saved into an XML file.

\vspace{0.02in}

\noindent \textbf{Implementation details}. The input content to our
scientific poster generation approach is also specified in an XML
file. This file specifies the structure and contents of a scientific
paper, including chapters, sections, paragraphs and graphical elements.
The other attributes such as caption and key words are also saved
in the corresponding content block. Note that the equation and formulas
are taken as normal texts since they can be written in latex format.
For graphical elements, we only save the width and height in the XML
file. In our experiment, sections and sub-sections corresponds to
panels and bullets respectively. We use TextRank to extract textual
content from the XML file. In order to give different importance of
different sections, we can set different extraction ratio for each
of them. This will result in important sections generating more content
and hence occupying bigger panels. For simplicity, this paper uses
equal important weights for all sections. The Bayesian Network Toolbox
(BNT)~\cite{Murphy02} is used for key parameters estimation and
sampling. For graphical element attributes inference, we generate
$1000$ samples by the likelihood-weighted sampling method \cite{weighted_sampling}
for Eq. \ref{eq:map_inference_function}. With the inferred metadata,
the final poster is generated in latex Beamerposter format with Lankton
theme. We will release all code upon paper acceptance.

\vspace{0.02in}

\noindent \textbf{Competitors and evaluation metrics }We compare several
baselines on different sections of our model to evaluate the methods
of attributes inference. Particularly, we compare ridge regression,
regression tree, support vector regression (SVR) with linear kernel
and RBF kernel respectively. And for graphical elements position ($h_{g}$)
inference, we regard it as a classification problem, then compare
the performance of our method with nearest neighbors classification
(KNN), decision tree, support vector classification (SVC) with linear
and RBF kernel. We employ the corresponding values for the original
(human) designed posters as the ground-truth. We split the dataset
into 80 pairs for training and validation, and the rest (5 pairs)
for testing.

\vspace{0.02in}

\noindent \textbf{Comparing with human designed posters.} We then
evaluate how well our approach facilitates scientific poster generation,
as compared to novice designers and the original poster (which is
designed by the author). We invite three second-year Phd students,
who are not familiar with our project, to hand design posters for
the test set. These three students work in computer vision and machine
learning and have not yet published any papers on these topics; hence
they are novices to research. Given the test set papers, we ask the
students to work together and design a poster for each paper.

\vspace{0.02in}

\noindent \textbf{Running time.} Our framework is very efficient in
term of running cost. Our experiments were done on a PC with an Intel
Xeon $3.6$ GHz CPU and $11.6$ GB RAM. Tab. \ref{tab:Effectiveness-of-each}
shows the average time we needed for each step. The total running
time is significantly less than the time people require to design
a good poster, it is also less than the time spent to generate the
posters made by three novices in Sec. \ref{subsec:Quantitative-Evaluation}.

\begin{table}
\centering{}%
\begin{tabular}{|c|c|r|}
\hline 
\multicolumn{2}{|c|}{\textbf{\small{}stage}} & \textbf{\small{}Average time}{\small{} }\tabularnewline
\hline 
\multicolumn{2}{|c|}{{\small{}Text extraction}} & {\small{}9.2362s }\tabularnewline
\hline 
\multirow{2}{*}{{\small{}Panel attributes inference} } & {\small{}learn }  & {\small{}0.33s }\tabularnewline
\cline{2-3} 
 & {\small{}infer }  & {\small{}0.004s }\tabularnewline
\hline 
\multicolumn{2}{|c|}{{\small{}Panel layout generation}} & {\small{}0.001s }\tabularnewline
\hline 
\multirow{2}{*}{{\small{}Composition within panel} } & {\small{}learn}  & {\small{}0.57s }\tabularnewline
\cline{2-3} 
 & {\small{}infer }  & {\small{}0.913 }\tabularnewline
\hline 
\end{tabular}\caption{\label{tab:Effectiveness-of-each}Running time of each step. }
\end{table}

\subsection{Quantitative Evaluation\label{subsec:Quantitative-Evaluation}}

\noindent \noindent \textbf{Effectiveness of attribute inferences.}
To validate the effectiveness of this step, our model is compared
against several state-of-the-art regression methods, including ridge
regression, regression tree, linear SVR and RBF-SVR.

The results are shown in Table \ref{tab: RMSE_Comparison}. We use
the panel attributes of original posters as the ground-truth and Root-Mean-Square
Error (RMSE) is computed for the inferred size and aspect ratio of
each panel. Specifically, we use the design of original poster as
the ground-truth and the RMSE is computed as,

\begin{equation}
RMSE=\sqrt{\sum_{i=1}^{n}(s_{p}-s_{p}^{'})/n}\label{eq:rmse}
\end{equation}
\noindent where $s_{p}$ represents the panel size of original panel,
and $s_{p}^{'}$ represent the panel size inferred by learning model;
$n$ indicates the total number of panels of all the posters. In Eq
(\ref{eq:rmse}), we use $s_{p}$ as an example; the RMSE for $r_{p}$
and $u_{g}$ can be calculated in the same way. 

To infer the panel size ($s_{p}$) and aspect ratio ( $r_{p}$) ,
we use the $t_{p}$ and $g_{p}$ as features. Comparing with all the
other methods, the RMSE of our method is only $0.71$ and $0.695$
respectively, which is lower than all the other methods. This shows
that our algorithm can better estimates the panel attributes than
other methods, due to our probabilistic graphical formulation that
effectively models the correlations and dependence among variables.

For graphical elements size ($u_{g}$) and horizontal position ($h_{g}$),
we use {$s_{p}$, $r_{p}$, $l_{p}$, $s_{g}$, $r_{g}$} as features
and our model is compared against all the other methods. RMSE and
accuracy is used to evaluate the performance of each method on $u_{g}$
and $h_{g}$, respectively. The accuracy is computed as

\begin{equation}
\begin{split}Accuracy & =\sum_{i=1}^{n}I(h_{g},h_{g}^{'})/n\\
I(h_{g},h_{g}^{'}) & =\begin{cases}
1,h_{g}=h_{g}^{'}\\
0,otherwise
\end{cases}
\end{split}
\end{equation}
\noindent where $h_{p}$ represents the horizontal position of original
panel, and $h_{p}^{'}$ represent the horizontal position inferred
by learning model. As shown in Tab. \ref{tab:The-accuracy-of_element}
and Tab. \ref{tab: RMSE_Comparison}, our results beat those of all
the other methods since design constraints are introduced in the inference
stage by Eq.(\ref{eq:map_inference_function}).

\begin{table}
\begin{centering}
\begin{tabular}{|c|c|c|c|c|c|}
\hline 
Attributes  & \rotatebox{90}{Our Method }  & \rotatebox{90}{Ridge Regression}  & \rotatebox{90}{Regression Tree }  & \rotatebox{90}{Linear-SVR }  & \rotatebox{90}{RBF-SVR} \tabularnewline
\hline 
\hline 
panel size ($s_{p}$)  & 0.071  & 0.075  & 0.090  & 0.073  & 0.120 \tabularnewline
\hline 
panel aspect ratio ($r_{p}$)  & 0.695  & 0.696  & 0.819  & 0.702  & 0.737\tabularnewline
\hline 
graphical element size ($u_{g}$)  & 0.0144  & 0.289  & 0.287  & 0.361  & 1.041\tabularnewline
\hline 
\end{tabular}
\par\end{centering}
\caption{\label{tab: RMSE_Comparison}Performance of comparing the methods
on panel size($s_{p}$), panel aspect ratio ($r_{p}$) and graphical
element size ($u_{g}$). RMSE is used as metric. Note that here we
only consider the relative size of each panel which is normalized
into $[0,1]$. The lower value, the better performance.}
\end{table}

\begin{table}
\begin{centering}
\begin{tabular}{|c|c|c|c|c|c|}
\hline 
Attributes  & \rotatebox{90}{Our Method}  & \rotatebox{90}{KNN}  & \rotatebox{90}{Decision Tree}  & \rotatebox{90}{Linear-SVC}  & \rotatebox{90}{RBF-SVC} \tabularnewline
\hline 
\hline 
horizontal position ( $h_{g}$)  & 88.9\%  & 66.7\%  & 66.7\%  & 72.2\%  & 72.2\% \tabularnewline
\hline 
\end{tabular}
\par\end{centering}
\caption{\label{tab:The-accuracy-of_element}The accuracy of predicting horizontal
position($h_{g}$). The higher value, the better performance.}

\end{table}

\begin{table}
\begin{centering}
\begin{tabular}{|c|c|c|c|c|}
\hline 
Metric  & Readability  & Informativeness & Aesthetics  & Avg.\tabularnewline
\hline 
\hline 
Our method  & 7.32  & 7.08  & 6.70  & 7.03\tabularnewline
\hline 
Posters by novices  & 6.82  & 6.80  & 6.58  & 6.73\tabularnewline
\hline 
Original posters  & 7.36  & 7.10  & 7.44  & 7.30\tabularnewline
\hline 
\end{tabular}
\par\end{centering}
\caption{\label{tab:User-Study-of}User study of different posters generated. }
\end{table}

\subsection{Qualitative User Study Evaluation}

\begin{figure}[!t]
\centering \begin{subfigure}[t]{0.22\textwidth} \frame{\includegraphics[width=1\textwidth]{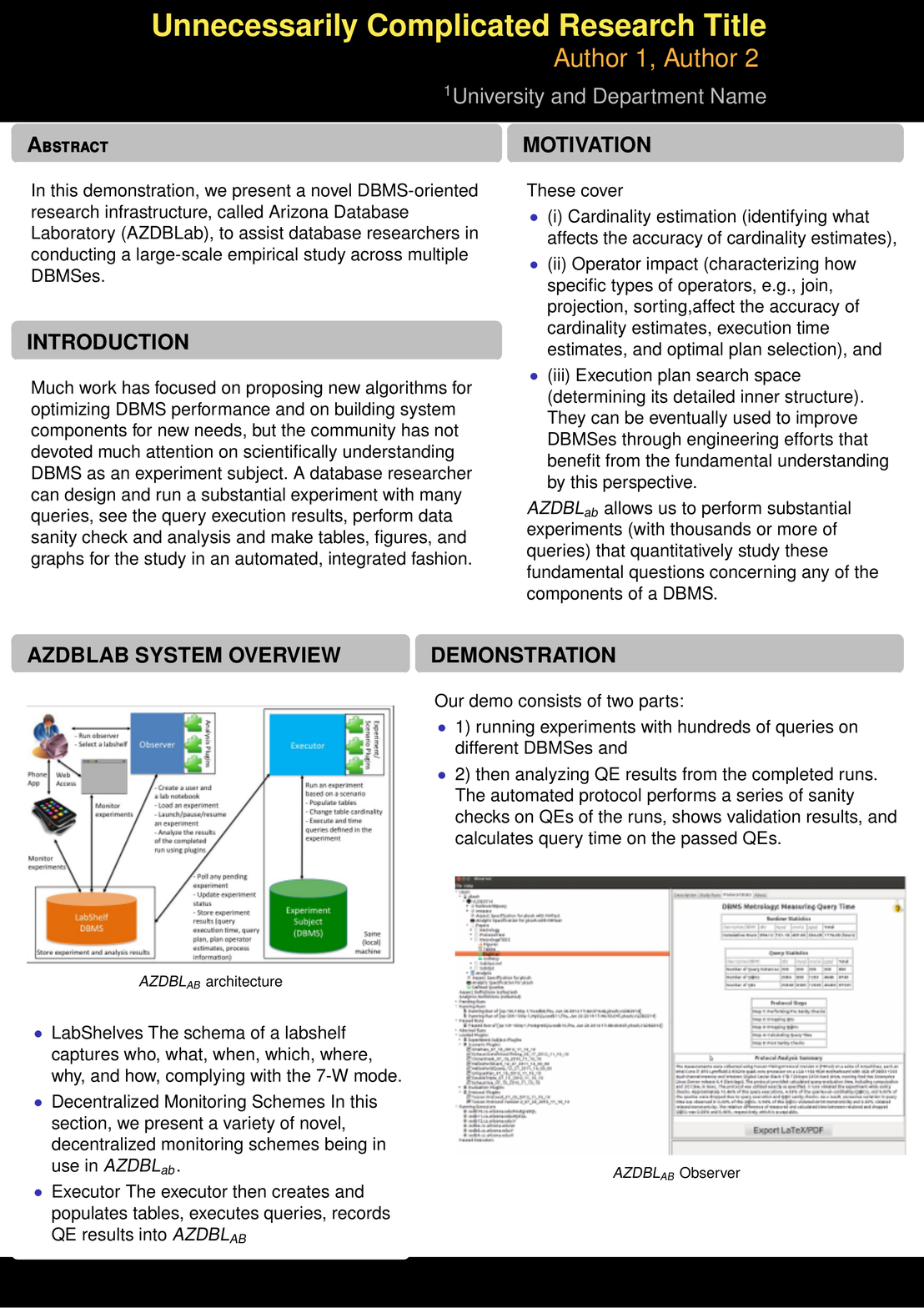}}
\protect{\caption{\label{fig:resa}Our method generates the poster of \cite{Suh2014VLDB}.}
} \end{subfigure} \hspace{20pt} \begin{subfigure}[t]{0.22\textwidth}
\frame{\includegraphics[width=1\textwidth]{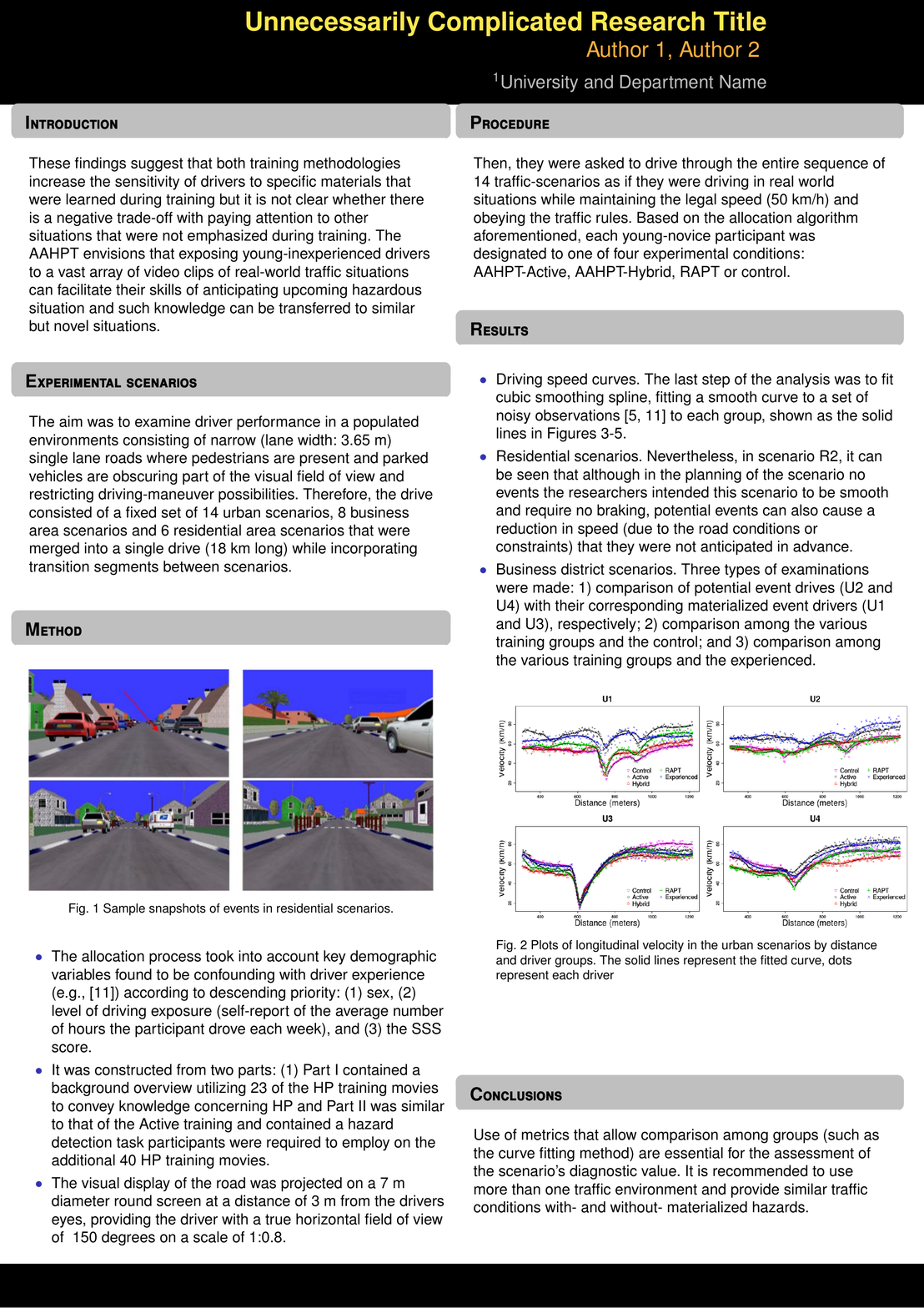}} \caption{ \label{fig:resb}Our method generates the poster of \cite{Oron2014ATS}.}
\end{subfigure} \protect\caption{\label{fig:res} Example of our results}
\end{figure}

\noindent \noindent \textbf{User study.} User study is employed to
compare our results with original posters and posters made by novices.
We invited 10 researchers (who are experts on the evaluated topic
and kept unknown to our projects) to evaluate these results on readability,
informativeness and aesthetics. Each researcher is sequentially shown
the three results generated (in randomized order) and asked to score
the results from $0-10$, where $0$, $5$ and $10$ indicate the
lowest, middle and highest scores of corresponding metrics. The final
results are averaged across subjects. Note that, since our method
mainly consider the layout of a poster, we provide novice designer
and our method with contents same as the original poster. We argue
that this is a more objective way to evaluate our method, because
both text extracted by TextRank and novice designer may not be as
good as the text in original poster, since it is summarized by the
author of the paper. And different contents would affect researchers
evaluation. 

In Table \ref{tab:User-Study-of}, on \emph{readability} and \emph{informativeness},
our result is comparable to the original poster; and it is significantly
better than posters made by novices. This validates the effectiveness
of our method. On one hand, the inferred panel attributes and generated
panel layout will save most valuable and important information. Besides,
composition within each panel that inferred by our method would give
proper emphasis on figures and tables, which may be overlooked by
novice designer. In contrast, our method is lower than the original
posters on aesthetics metric (yet, still higher than those from novices).
This is reasonable, since aesthetics is a relatively subjective metric
and it generally needs to involve lots of human interactions. Human
designers can adjust the the poster layout via lots of latex commands
again and again. In general, it is an open problem to generate more
aesthetic posters from papers.

\begin{figure*}[thb!]
\centering \begin{subfigure}[b]{0.29\textwidth} \frame{\includegraphics[width=1\textwidth]{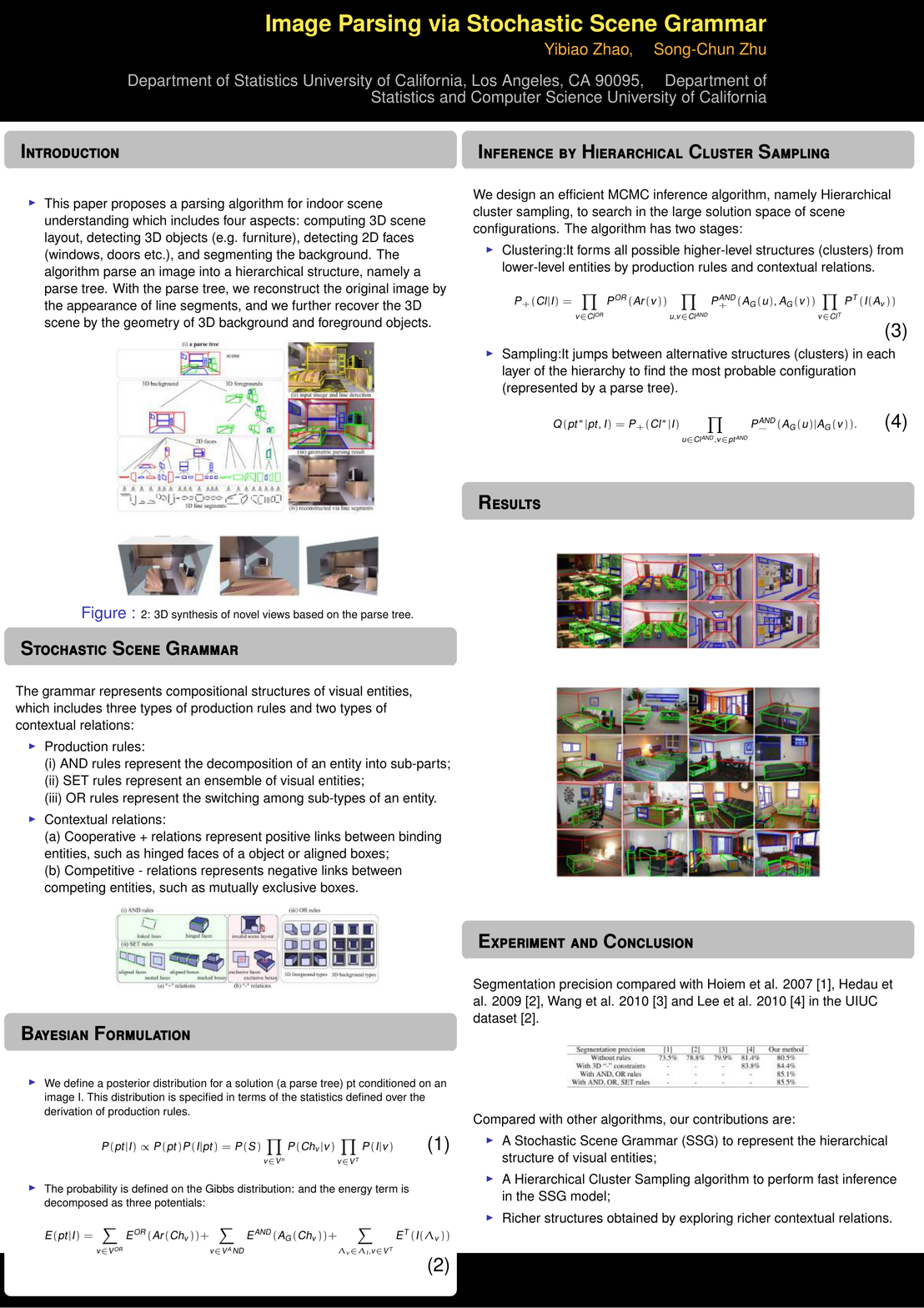}}
\caption{\label{fig:novice_res1}Designed by novice}
\end{subfigure} 
\begin{subfigure}[b]{0.29\textwidth} \frame{\includegraphics[width=1\textwidth]{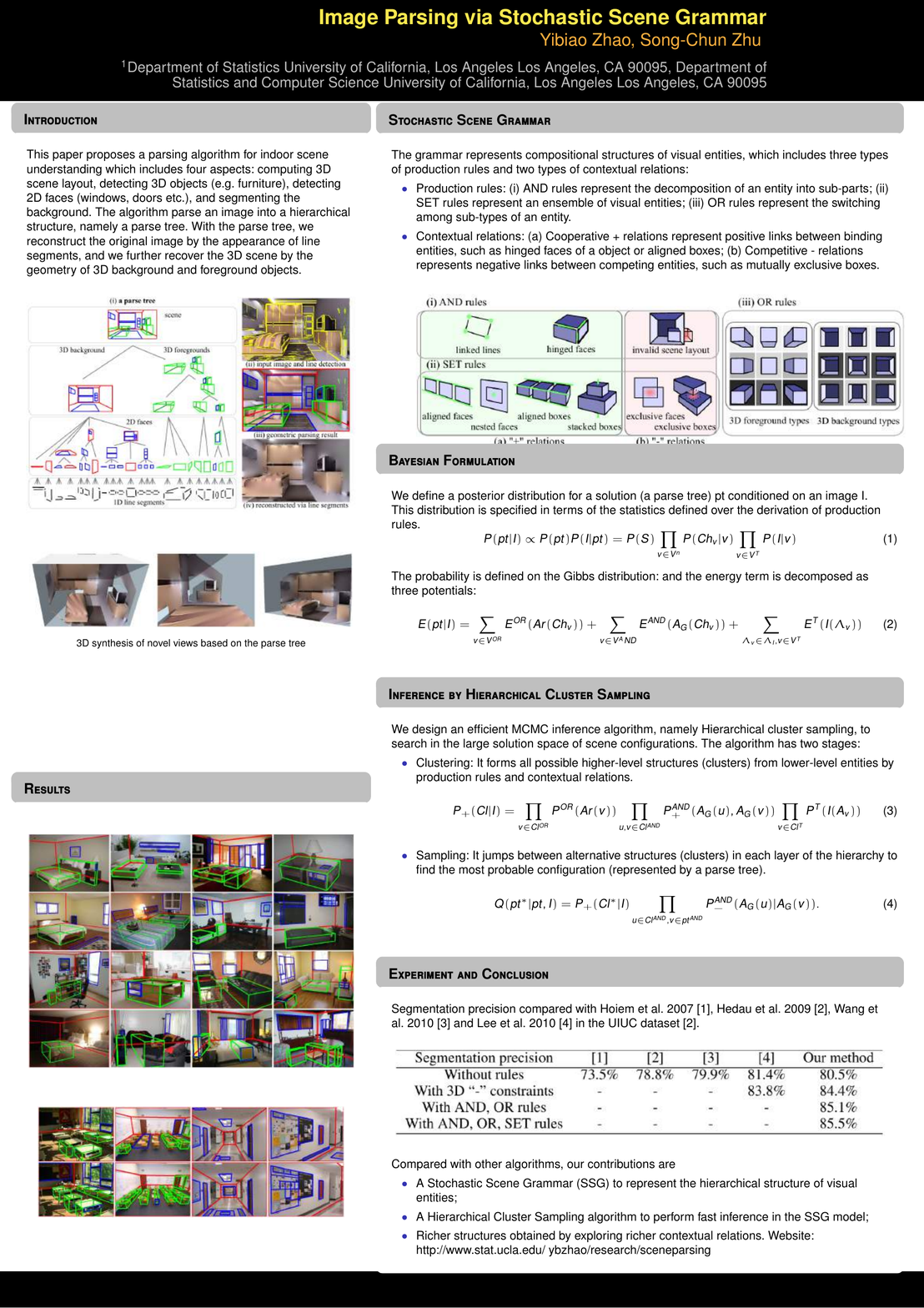}}
\caption{\label{fig:our_res1}Our result}
\end{subfigure} \vspace{1em}
 \begin{subfigure}[b]{0.308\textwidth} \frame{\includegraphics[width=1\textwidth]{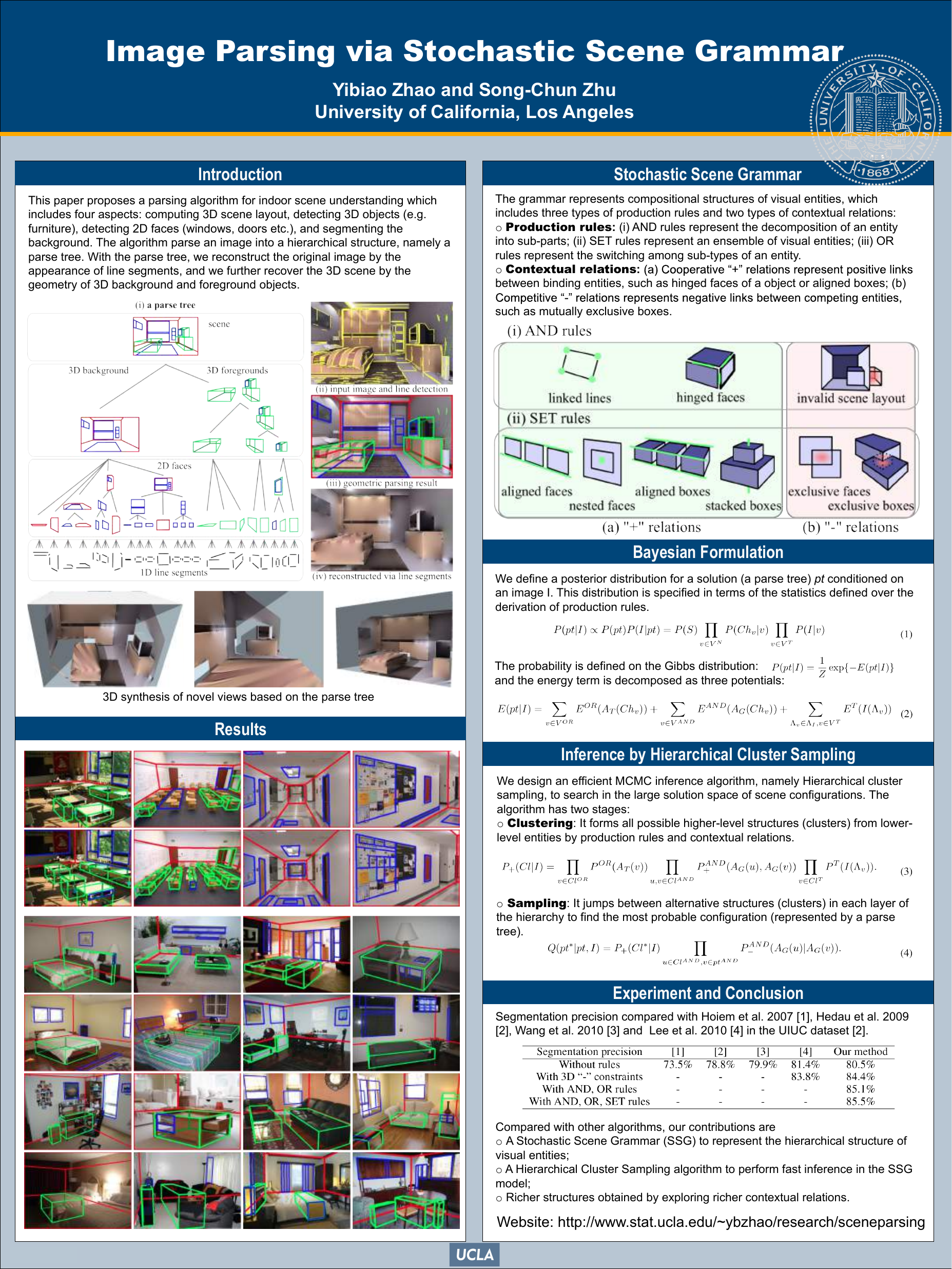}}
\caption{\label{fig:original_res1}Original poster\cite{Yibiao2011NIPS}}
\end{subfigure} \centering \begin{subfigure}[b]{0.30\textwidth}
\frame{\includegraphics[width=1\textwidth]{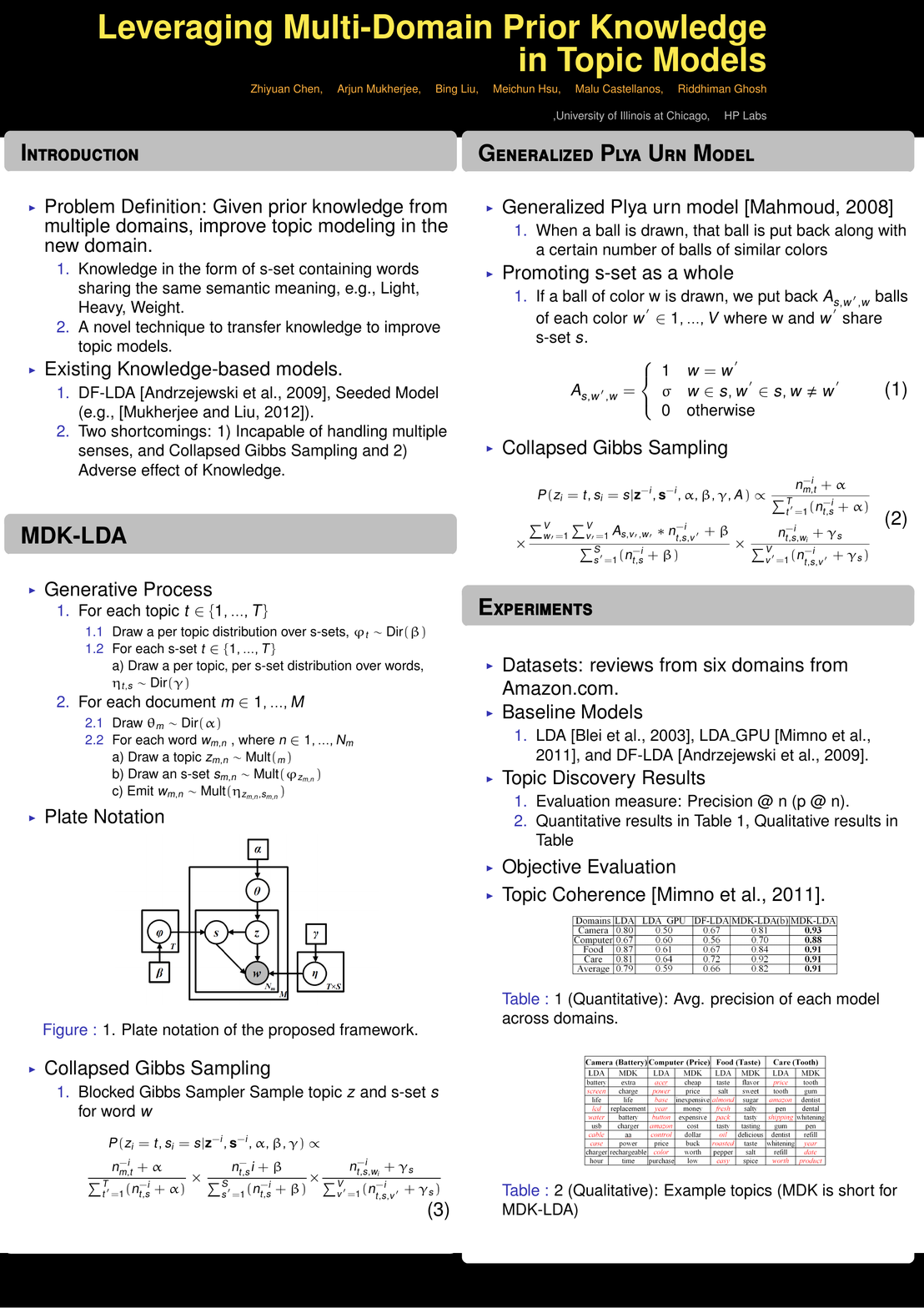}}
\caption{\label{fig:novice_res2}Designed by novice}
\end{subfigure} 
\begin{subfigure}[b]{0.30\textwidth} \frame{\includegraphics[width=1\textwidth]{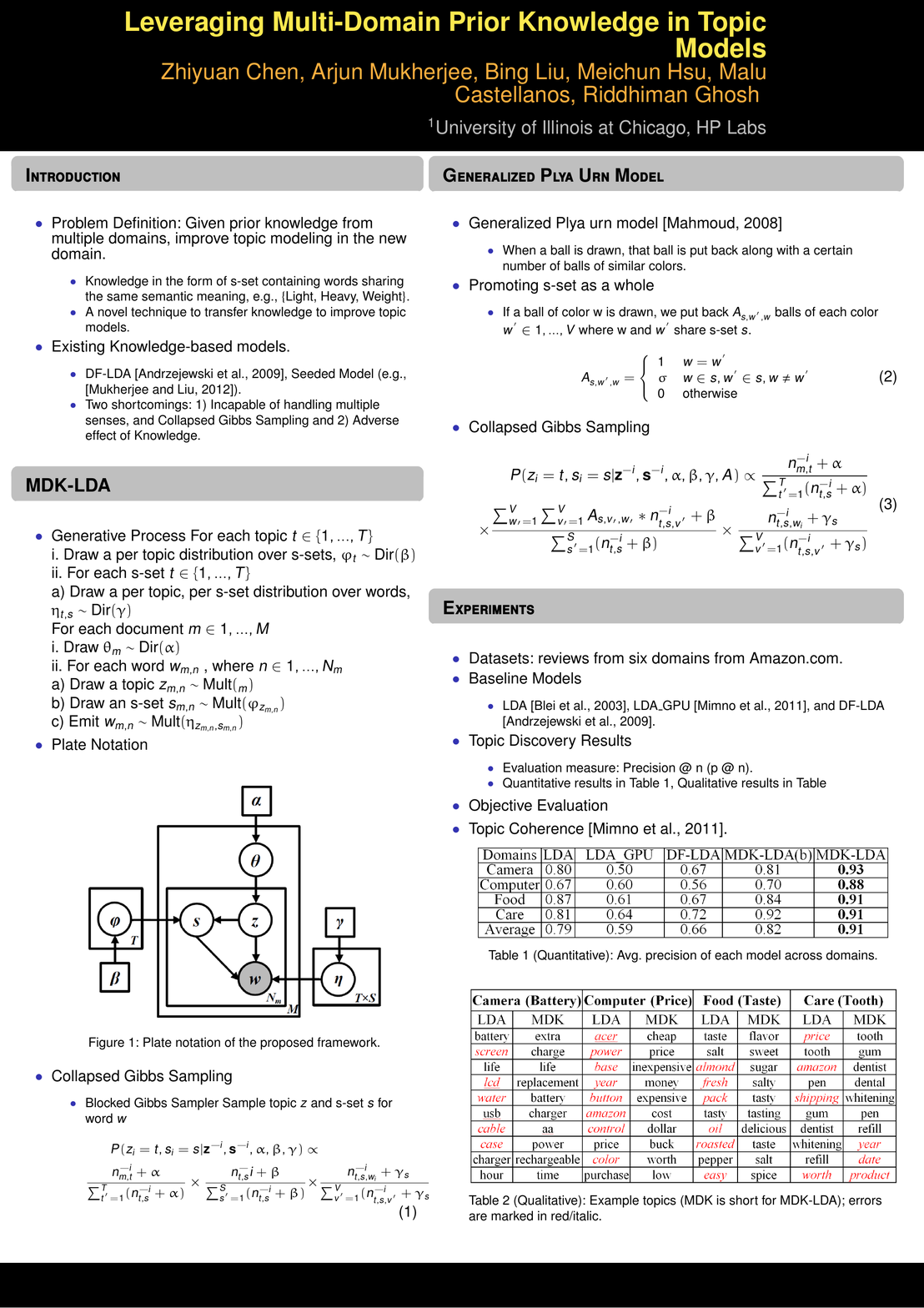}}
\caption{\label{fig:our_res2}Our result}
\end{subfigure} \vspace{1em}
 \begin{subfigure}[b]{0.30\textwidth} \frame{\includegraphics[width=1\textwidth]{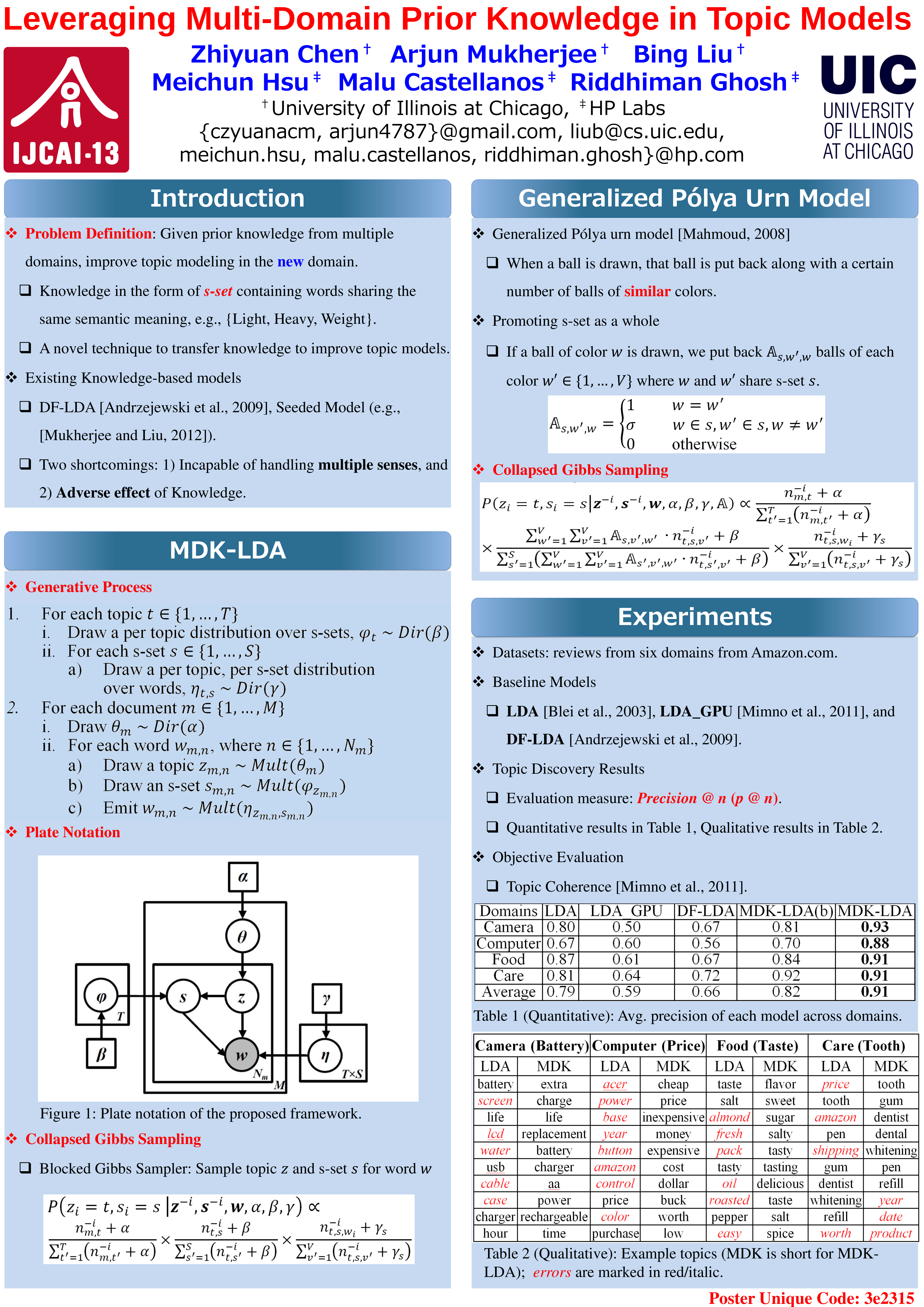}}
\caption{\label{fig:original_res2}Original poster\cite{Zhi2013IJCAI}}
\end{subfigure} \caption{\label{fig:results}Results generated by different ways}
\label{fig:results} 
\end{figure*}

\noindent \noindent \textbf{Qualitative Evaluation of Three Methods.
}We qualitatively compare our results (Figure \ref{fig:our_res1}
and Figure \ref{fig:our_res2}) with the posters from novices (Figure
\ref{fig:novice_res1} and Figure \ref{fig:novice_res2}) and the
original posters (third blob in Figure \ref{fig:original_res1} and
Figure \ref{fig:original_res2}). All of them are for the same paper
and with same contents. 

It is interesting to show that if compared with the panel layout of
original poster, our panel layout looks more similar to the original
one than the one by novices. This is due to, firstly, the Paper-Poster
dataset has a relatively similar graphical design with high quality,
and secondly, our split and panel layout algorithm works well to simulate
the way how people design posters. In the first row of Figure \ref{fig:results},
we can see that in order to arrange contents in the poster aesthetically,
the order of each panel is rearranged in poster from the novice designer
(Figure \ref{fig:novice_res1}), this would affect the readability
of a poster. The second row of Figure \ref{fig:results} shows that,
compared with novice designer, our method also achieve good performance
on attributes inference for graphical elements. The size of graphical
elements inferred by our method seems similar to the original poster.
In contrast, the poster designed by novices in Figure \ref{fig:novice_res2}
lose emphasis on figures in order to keep the content fit each panel. 

\subsection{Qualitative Evaluation by Design Principles}

\noindent We further qualitatively evaluate our results (Figure \ref{fig:res})
by the general graphical design principles \cite{odonovan2014}, i.e.,
\emph{flow}, \emph{alignment},and \emph{overlap and boundaries}.

\vspace{0.02in}

\noindent \textbf{Flow.} It is essential for a scientific poster to
present information in a clear read-order, i.e. readability. People
always read a scientific poster from left to right and from top to
bottom. Since Algorithm \ref{alg:panel_gen} recursively splits the
page of poster into \emph{left, right} or \emph{top, bottom}, the
panel layout we generate ensure that the read-order matches the section
order of original paper. Within each panel, our algorithm also sequentially
organizes contents which also follow the section order of original
paper and this improves the readability.

\vspace{0.02in}

\noindent \textbf{Alignment}. Compared with the complex alignment
constraint in \cite{odonovan2014}, our formulation is much simpler
and uses an enumeration variable to indicate the horizontal position
of graphical elements $h_{g}$. This simplification does not spoil
our results which still have reasonable alignment as illustrated in
Figure \ref{fig:res} and quantitatively evaluated by three metrics
in Table \ref{tab:User-Study-of}.

\vspace{0.02in}

\noindent \textbf{Overlap and boundaries}. Overlapped panels will
make the poster less readable and less aesthetic. To avoid this, our
approach (1) recursively splits the page for panel layout; (2) sequentially
arranges the panels; (3) Design constraint in incorporated into our
Bayesian network (Eq.\ref{eq:map_inference_function}) to penalize
the cases of overlapping between graphical elements and panel boundaries.
As a result, our algorithm can achieve reasonable results without
significant overlapping and/or crossing boundaries. Similar to the
manually created poster \textendash{} Figure \ref{fig:results}(c),
our result (i.e., Figure \ref{fig:results}(b)) does not have significantly
overlapped panels and/or boundaries.

\section{Conclusion and Future Work}

\label{sec:conclusion_and_future_work} Automatic tools for scientific
poster generation are important for poster designers. Designers can
save a lot of time with these kinds of tools. Design is a hard work,
especially for scientific posters, which require careful consideration
of both utility and aesthetics. Abstract principles about scientific
poster design can not help designers directly. In contrast, we propose
an approach to learn design patterns from existing examples, and this
approach can be used as an assistant tool of scientific poster generation
to aid the designers.

As the future work, our framework can be also applicable to directly
learn the general design patterns such as the web-page design, and
single-page graphical design, if given the corresponding layout styles.
Currently, we do not consider font types of posters which will be
addressed in future. \bibliographystyle{abbrv}
\bibliography{references}
 
\end{document}